\documentclass[10pt,letterpaper]{article}

\usepackage{cogsci}
\usepackage{pslatex}
\usepackage[hidelinks]{hyperref}

\usepackage{apacite}

\usepackage{graphicx}
\usepackage{xcolor}
\usepackage{algorithm}
\usepackage{algpseudocode}
\usepackage{tabularx}
\usepackage{tikz-cd}
\usepackage{qtree}
\usepackage{tikz-qtree-compat}
\usepackage{subfig}
\usepackage{dsfont}
\usepackage{amsmath} 
\usepackage{booktabs}
\usepackage{dsfont}

\usepackage{multirow}
\usepackage{subcaption}
\usepackage{booktabs}  
 \usepackage{longtable} 
 \usepackage{enumitem}

\usepackage{tikz}

\usetikzlibrary{arrows.meta, positioning, fit}

\usepackage{todonotes}
\cogscifinalcopy   

\title{A computational operationalisation of competing maturational theories of syntactic development via statistical grammar induction}

\author{
\large
\begin{tabular}{ccc}
\textbf{Mila Marcheva-Nash} & \textbf{Suchir Salhan} & \textbf{Weiwei Sun} \\
{\normalfont\ttfamily mmm67@cam.ac.uk} &
{\normalfont\ttfamily sas2450@cam.ac.uk} &
{\normalfont\ttfamily ws390@cam.ac.uk}
\end{tabular}
\\[0.5em]
Department of Computer Science \& Technology, University of Cambridge
}

\begin{document}

\maketitle

\begin{abstract}
This paper is concerned with what intermediate syntactic categories children acquire during first language development, and in what order. Maturational theories make different predictions. Bottom-up accounts (\textsc{Growing}) propose that lexical and inflectional structure emerges first, while inward accounts (\textsc{Inward}) predict early access to discourse-related categories.
We computationally operationalise these hypotheses of staged syntactic emergence using statistical grammar induction, asking what each proposed ordering makes learnable when input and learning algorithm are held constant.
Our framework makes category acquisition explicit and allows us to explore how different maturational orderings shape the structure that can be learned under identical conditions. Based on this operationalisation, the \textsc{Growing} account significantly outperforms the \textsc{Inward} account across three evaluation metrics. 
\end{abstract}




\section{Introduction}

A central question in first language acquisition (FLA) is how children develop an adult-like grammatical system \cite{brown1973development}. Linguistic theories differ on whether grammatical categories are predetermined by biology or emerge gradually from experience. Continuity approaches within the generative tradition assume that all categories are innate and available from birth, consistent with the idea of Universal Grammar (UG) \cite{Pinker1984, BoserLustSantelmannWhitman1992, ClahsenEisenbeissVainikka1994}. Maturational accounts assume that certain syntactic categories are innate but become accessible only at specific points in development, shaping the order of acquisition \cite{Radford1988, Radford1990, Guasti1993}. By contrast, the emergentist perspective emphasises that categories emerge from patterns in the input under cognitive constraints \cite{OGrady2008, boschbiberauer2024bucld}. Functionalist, usage-based, and constructivist theories are in line with emergentism \cite{BatesMacWhinney1987, TOMASELLO2005, LievenTomasello2008, Ambridge2015, LIEVEN_2016, BEHRENS2021}.
What is common across maturational accounts and emergentist accounts is the rejection of continuity, i.e. both theories agree that adult grammatical competence is not available from the outset \cite{Tomasello2000, boschgran2025}. What maturation and emergentism disagree on is the source for the staged development: under maturation, the order in which the categories appear is innately encoded, whereas under emergentism, the order is determined by the interplay of input and cognitive constraints.

Within the maturational tradition theories differ sharply on which categories appear first. Bottom-up proposals, such as the \textsc{Growing} Trees Hypothesis \cite{Friedmann2021}, suggest that lexical and inflectional categories (\texttt{N}, \texttt{V}, \texttt{T}) emerge first, followed by discourse-related functional categories. Inward maturation proposals, like the \textsc{Inward} Growing Spine Hypothesis \cite{Heim2025}, predict early access to higher discourse-related categories, with lower categories appearing later. These hypotheses formalise patterns observed in corpora of child and child-directed speech. 
Computational modelling can complement traditional methods by allowing us to explore whether the proposed developmental orderings make different grammatical structures recoverable from the same input.

In this paper, we introduce a computational framework for testing competing theories of staged syntactic category acquisition under developmentally plausible constraints. In grammar induction, the statistical learner receives strings as input and infers explicit hierarchical structures, a grammar, making the order of category acquisition observable rather than stipulated \cite{johnson2013language}.
By controlling the order in which categories become accessible, we simulate the developmental trajectories predicted by different maturational theories, while holding constant the input and the learning algorithm. 
Comparing the resulting induced grammars across controlled conditions reveals how different maturational orderings constrain what is learnable from identical input and under identical conditions.



We present an experimental suite to compare two maturational hypotheses, \textsc{Growing} and \textsc{Inward}. We use morphemically tokenised child-directed speech as developmentally plausible input, reflecting linguistic units to which children are known to be sensitive \cite{pearl-sprouse-2013, marcheva-etal-2025-profiling}. 
Across a wide range of conditions, we find empirical advantages for staged category acquisition, particularly under the \textsc{Growing} curriculum, relative to Continuity approaches. While both \textsc{Growing} and \textsc{Inward} curricula converge to comparable global performance, \textsc{Growing} significantly outperforms \textsc{Inward} on all queried metrics (F1, Jensen-Shannon divergence, and child speech sentence log-likelihood). At the level of the learning process, \textsc{Growing} favours earlier stabilisation of phrasal structure, whilst \textsc{Inward} yields lower divergence for certain clause-level categories. 
Together, these results align with the qualitative predictions of the corresponding maturational accounts and demonstrate the utility of our method for systematically exploring alternative trajectories of staged syntactic development. The framework and the tested staged grammars are available on GitHub\footnote{\url{https://github.com/milamarcheva/maturational_grammar_induction}}. While we pilot the framework on two maturational hypotheses, the framework could be extended to emergentist hypotheses, given appropriate staged curricula are developed.

\begin{figure}[h!]
\centering
\resizebox{1\linewidth}{!}{%
\begin{tikzpicture}[
    font=\footnotesize,
    >=Latex,
    block/.style={
        draw,
        rounded corners=5pt,
        align=left,
        text width=9.6cm,
        inner sep=6pt,
        fill=white
    },
    subblock/.style={
        draw,
        rounded corners=4pt,
        align=left,
        text width=8.6cm,
        inner sep=4.5pt,
        fill=gray!7
    },
    output/.style={
        draw,
        rounded corners=4pt,
        align=left,
        text width=8.6cm,
        inner sep=4.5pt,
        fill=cyan!12
    },
    eval/.style={
        draw,
        dashed,
        rounded corners=4pt,
        align=left,
        text width=8.6cm,
        inner sep=4.5pt,
        fill=orange!16
    },
    arrow/.style={->, thick},
    looparrow/.style={->, thick, gray!75}
]


\node[block] (prep) {
    {\bfseries 1. Prepare staged grammar}\\[2pt]
    \textsc{Oracle} PCFG: symbolic component (rules) + probabilistic component\\
    Split symbolic component into maturational stages
};


\node[block, below=0.38cm of prep] (initial) {
    {\bfseries 2. Initial stage}\\[3pt]

    \begin{tikzpicture}[
        node distance=0.1cm,
        every node/.style={font=\footnotesize}
    ]
        \node[subblock] (initvb) {
            Set inter-stage transfer parameters, \(s_p, s_\ell, \eta\);
            Run VB on first stage
        };

        \node[output, below=of initvb] (initout) {
            \textbf{Output:} estimated probabilities for first-stage rules 
        };

        \node[eval, below=of initout] (initeval) {
            \textbf{Evaluation:} $G_1$, grammar up to stage 1
        };
    \end{tikzpicture}
};


\node[block, below=0.38cm of initial] (loop) {
    {\bfseries 3. Repeat for each following stage}\\[3pt]

    \begin{tikzpicture}[
        node distance=0.1cm,
        every node/.style={font=\footnotesize}
    ]
        \node[subblock] (add) {
            Add newly available symbolic rules
        };

        \node[subblock, below=of add] (priors) {
            Update priors:
            existing rules \(P_k \mapsto \alpha_{k+1}\);
            new rules receive mass via \(\eta\)
        };

        \node[subblock, below=of priors] (vb) {
            Run VB re-estimation for current staged grammar
        };

        \node[output, below=of vb] (outk) {
            \textbf{Output:} probabilities for all rules available so far
        };

        \node[eval, below=of outk] (evalk) {
            \textbf{Eval.:} $G_i$, grammar up to stage $i$
        };

        \draw[looparrow]
            (evalk.east) .. controls +(0.9,0.0) and +(0.9,0.0) ..
            node[pos=0.6, xshift=-0.25cm, align=right, font=\scriptsize, gray!80]
            {next\\stage}
            (add.east);
    \end{tikzpicture}
};


\node[block, below=0.38cm of loop] (final) {
    {\bfseries 4. After all stages}\\[3pt]

    \begin{tikzpicture}[
        node distance=0.1cm,
        every node/.style={font=\footnotesize}
    ]
        \node[output] (finalout) {
            \textbf{Output:} probabilities for the entire symbolic grammar
        };

        \node[eval, below=of finalout] (finaleval) {
            \textbf{Eval.:} G, final grammar
        };
    \end{tikzpicture}
};

\draw[arrow] (prep) -- (initial);
\draw[arrow] (initial) -- (loop);
\draw[arrow] (loop) -- (final);

\end{tikzpicture}%
}

\caption{Pipeline for statistical learner modelling staged syntactic development, explained in detail in \hyperref[sec:methodology]{\textbf{Methodology}}.}
\label{fig:pipeline}
\end{figure}
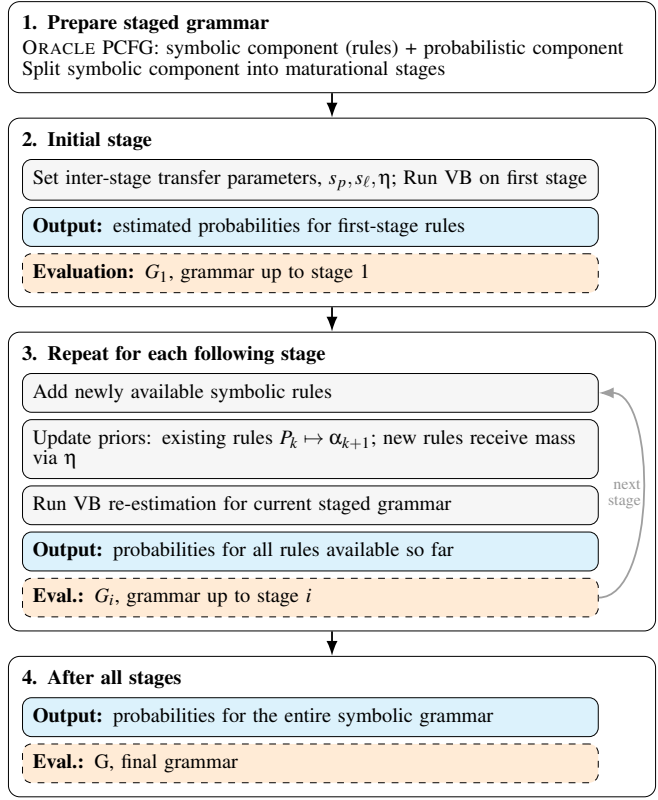

\section{Background}
\label{sec:back}

\subsection{Syntactic development}

A long tradition in FLA characterises syntactic development as staged, following a trajectory of single-word utterance, then unmarked two-word phrases, and ultimately phrases marked with increasingly complex morphosyntactic structure \cite{brown1973development}.
Generative accounts of syntactic acquisition centre on when and how functional categories are acquired. There are two main schools of thought, with regards to when functional categories, assumed to be part of Universal Grammar (UG), become available. \textbf{Continuity}~\cite{Westergaard2009} poses that all information in UG is available from the start. Thus, the functional structure of children's initial grammar is not significantly different from adult's grammars. 
\textbf{Maturation}~\cite{Friedmann2021, Heim2025} poses that grammatical categories and principles from UG are not fully available to children initially, but instead are incrementally accessed. Thus, UG specifies not only the hierarchical structure and functional categories, but the order in which they become available to the acquirer, too. Under maturation the UG-given hierarchical structure is fixed, but children gradually gain more access to parts of the structure. 
All of the above hypotheses assume that functional categories have fixed granularity. Thus, for completion we must also mention neo-emergentism, a hypothesis of syntactic acquisition which poses that increasing (flexible) granularity of the functional categories is a key aspect of syntactic acquisition \cite{biberauer_roberts2015, boschbiberauer2024bucld, boschgran2025}.

With the distinction of order of acquisition, maturation can be further broken down into bottom-up and inward orders (w.r.t. to the UG predefined spine). Bottom-up maturation poses that the categories which become available first are the ones closer to the leaves. The most recent bottom up approach, which also incorporates cartography, is the \textbf{\textsc{Growing} Trees Hypothesis} \cite{Friedmann2021}. It distinguishes between 3 stages: in stage 1 only IP/TP and VP are available (allowing for inflection and A(rgument)-movement);  in stage 2 the lower left-periphery (e.g. allowing for wh- questions) becomes available; finally in stage 3 the entire cartographic hierarchy becomes available, including topicalisations and embeddings. Inward maturation, of which an example is the \textbf{\textsc{Inward} Growing Spine Hypothesis} \cite{Heim2025}, postulates the early development of CP, which appears at the middle of the cartographic spine.





\begin{table*}[t]
\centering
\footnotesize
\renewcommand{\arraystretch}{1.25}
\begin{tabular}{p{0.09\textwidth} p{0.25\textwidth} p{0.66\textwidth}}
\toprule
\textbf{Stage} & \textbf{Description} & \textbf{Penn Treebank tags} \\
\midrule

\textsc{base}\newline\textsc{Growing}
& Simple NPs and VPs, \newline no other phrasal material
& \textbf{NTs:} ROOT, S, FRAG, NP, VP \newline
  \textbf{PTs:} NN, NNP, PRP, PRP\$, VB 
\\
\hline

\textsc{base}\newline\textsc{Inward}
& Discourse-level combinations\newline (e.g. \texttt{INTJ $\rightarrow$ INTJ INTJ})
& \textbf{PTs:} NN, PRP, PRP\$, VB, UH \newline
\textbf{NTs} ROOT, S, FRAG, INTJ 
\\\hline

\textsc{VP}
& Predicate and modifier structure
& \textbf{NTs:} ADJP, ADVP, PP, NNPS, NNS \newline
  \textbf{PTs:} DT, PDT, POS, NNS, CD, JJ, JJR, JJS, RB, RBR, RBS, IN, RP, VBG, VBN, NOT, DIV 
\\
\hline

\textsc{TP}
& Finite/tense and auxiliary marking
& \textbf{PTs/NTs:} VBD, VBP, VBZ, AUX, COP, MD, ASP, T, PRS 
\\\hline

\textsc{CP}
& Complementisers and operators
& \textbf{PTs:} COMP, CC, WP, WP\$, WRB, WDT \newline
  \textbf{NTs:} SBAR, SBARQ, SINV, SQ, WHADJP, WHADVP, WHNP 
  \\\hline
\textsc{INTJ}
& Interjections
& \textbf{PTs:} UH \newline \textbf{NTs:} INTJ \\
\end{tabular}
\caption{Syntactic stages are defined as cumulative sets of Penn Treebank categories. Note that in the original PTB some tags are PTs, but here are NTs due to the morphemic tokenisation (e.g. NNS, VBG).}
\label{tab:syntactic-stages}
\end{table*}

\subsection{Grammar induction}
Grammar induction (GI) is the process of learning the hierarchical structure which is latent in language data. 
A grammar is composed of a \textbf{symbolic component} (the rules) and a \textbf{probabilistic component} (the probabilities assigned to the rules). A subtype of GI is grammar reestimation \cite{pereira-schabes-1992-inside-outside}, where the symbolic component is provided, and the probabilistic component needs to be learned.  
Foundational work in GI relies on statistical methods \cite{klein-manning-2004-corpus, johnson-etal-2007-bayesian, carroll1992, bisk-hockenmaier-2012-induction, clark-schuler-2023-categorial}.
In recent years neural grammar induction has been put forward as a method which induces grammars with unprecedented F1 scores from raw data \cite{kim-etal-2019-compound}. However, neural methods are less interpretable than the foundational statistical models. Thus, for this paper we rely on statistical GI, specifically grammar reestimation, with a PCFG.

Using GI as an approximation for language acquisition, and specifically syntax acquisition, is well motivated in the literature \cite{johnson2013language}; examples include i.a. \citeA{buttery-2004-quantitative, Perfors2011, kwiatkowski-etal-2012-probabilistic}).
A key advantage of GI over traditional linguistic analyses, which often target isolated phenomena, is that it provides a unified account of all sentences in a corpus within a single model. Note that GI remains an exploratory tool and does not capture the full complexity of first language acquisition, as it abstracts away from non-language cues \cite{Dupre2024}.


\subsubsection{Probabilistic Context-Free Grammar (PCFG)}
A PCFG is defined as \(G=(NT,\Sigma,R,S,F)\), where \(NT\) is the set of non-terminals, \(\Sigma\) the vocabulary, \(R\) the rules, \(S\) the start symbol, and \(F\) the rule-probability function. We distinguish production rules from lexicalisations, since the former are the main object of staged syntactic access.

\subsubsection{Variational Bayes (VB)}
\label{sec:vb}
Statistical GI standardly uses one of two estimation methods: Expectation Maximisation~(EM)~\cite{klein-manning-2002-generative} or Variational Bayes~(VB)~\cite{Kurihara2006}. EM is a frequentist procedure that estimates rule probabilities solely from expected counts in the data, whereas VB is a Bayesian extension of EM that introduces prior distributions over rule probabilities, most commonly Dirichlet priors \cite{johnson-etal-2007-bayesian, liang-etal-2007-infinite}. 
In VB, each grammar rule is associated with a pseudocount parameter ($\alpha$) that influences learning at every iteration, encoding prior confidence in a rule. Higher values bias the model toward retaining a rule, while lower values allow unsupported rules to shrink in probability.  Pseudocounts allow probabilities learned in earlier stages to be carried forward as priors, while still permitting new rules to be learned.

\section{Methodology}
\label{sec:methodology}
\autoref{fig:pipeline} illustrates our computational operationalisation of staged syntactic development: maturational hypotheses are translated into curricula comprised of rule subsets of a symbolic grammar, and acquisition is approximated as Bayesian grammar induction.

\subsection{Acquisition-inspired curricula}
\label{sec:maturation}
In order to translate the theoretical claims about maturation and the timing of CP into a curriculum-based grammar induction problem, we define two curricula approximating the incremental access to syntactic categories stipulated by \textsc{Growing} and \textsc{Inward} maturation hypotheses. Previously
\citeA{salhan-etal-2024-less} have approached this problem using universal part-of-speech~(UPOS) tags \cite{petrov-etal-2012-universal}, however, we instead rely on the Penn Treebank (PTB) tagset \cite{PTB2}, which is more detailed. The PTB tagset encodes both phrase type (e.g. VP, NP, PP, INTJ) and the presence of functional material (e.g. AUX, MD, TO, COP, complementisers, wh-phrases), which allows us to define more fine-grained cognitively-inspired curricula. Stages are defined cumulatively over Penn Treebank tags in \autoref{tab:syntactic-stages}. Note that the stages we define are an approximation of the maturational theories, and alternative definitions of the stages can be substituted within the same pipeline.

The two maturational curricula, constructed by ordering the stages from \autoref{tab:syntactic-stages} are listed below, as well as an explicit statement of the \textsc{Continuity} condition:

\begin{itemize}[noitemsep,nolistsep]
    \item \textbf{\textsc{Growing}}: \textsc{baseGrowing}, \textsc{VP}, \textsc{TP}, \textsc{CP}, \textsc{INTJ}
    \item \textbf{\textsc{Inward}}: \textsc{baseInward}, \textsc{baseGrowing}, \textsc{CP}, \textsc{TP}, \textsc{VP}
    \item \textbf{\textsc{Continuity}}: all rules are available from the start
\end{itemize}

\subsection{Maturational syntactic development via VB}
Maturational syntactic development poses that syntactic knowledge becomes accessible incrementally in distinct stages (refer to \autoref{tab:syntactic-stages} for a PTB approximation of these stages).
Using VB we can perform learning in stages, where increasingly more rules become available. As new rules are being introduced at every stage, pseudocounts allow probabilities learned in earlier stages to be carried forward as priors, while still permitting new rules to be learned. 


We formulate the pseudocount of existing and new rules in Eq.~\ref{eq:existing}, which allows information to be carried across stages as the learning space expands. After the completion of stage $k$, the posterior mean rule probabilities $\mathds{P}^{k}$ are converted into a Dirichlet prior vector $\alpha^{k+1}$ for production rules $x\in X$ in stage $k{+}1$ according to:
\begin{equation}
\text{$\alpha$}^{k+1}_{i,x} = 
\begin{cases} 
N^k \, s_p \, {p}^{k}_i + 0.1, & 1 \leq i \leq N^k 
\\
\frac{N^k \, s_p \, \text{$\eta$}}{X^{k+1}-X^k} + 0.1, &  N^k < i \leq N^{k+1} \\
\end{cases} 
\label{eq:existing}
\end{equation}

where $N^k$ is the number of sentences successfully parsed at stage $k$, $s_p$ is a scaling parameter for production rule priors carried forward from the previous stage, and $\eta$ is a dial for how much probabilistic mass can be allotted to newly introduced rules in a stage. $X^{k}$ is the number of rules which share the same NT on the LHS as $x$ in stage $k$, so the mass for new rules is distributed across the newly available expansions for that NT. Lexicalisation priors are treated analogously, using a separate scaling parameter $s_\ell$.\footnote{Treating lexicalisations and productions with separate pseudocount values allows us to focus specifically on the production rules as they are more pertinent to the topic of syntactic development.}
Thus, $s_p$ and $s_\ell$ control the degree of inter-stage memory, while $\eta$ controls how readily the learner assigns probability mass to newly available rules. When $s=0$, learning from the previous stage is ignored, whereas larger values increasingly bias the learner toward retaining previously learned distributions.

\begin{figure*}[ht!]
    \centering
    \includegraphics[width=0.9\linewidth]{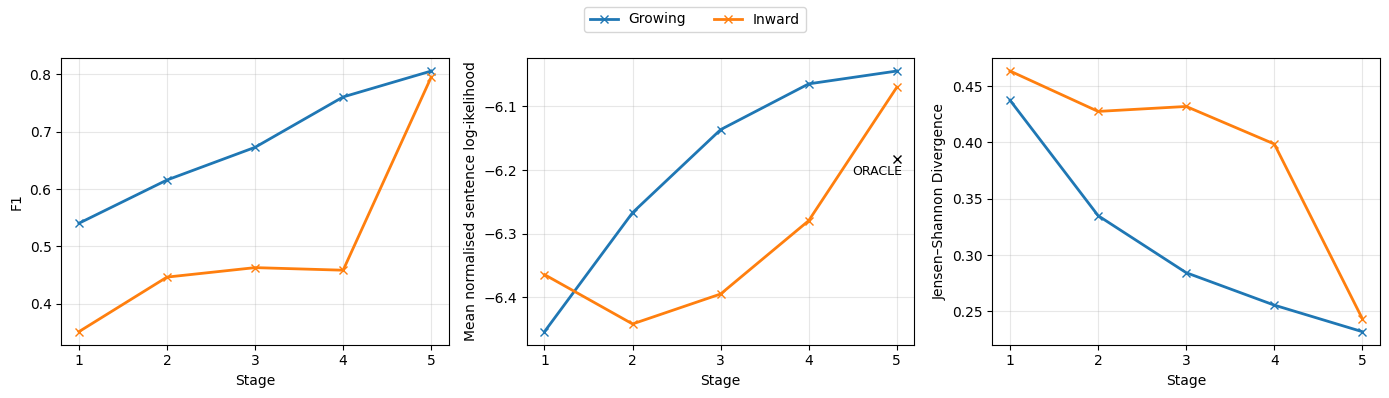}
    \caption{\textsc{\textsc{Growing}} vs \textsc{\textsc{Inward}} learning progression across the 5 learning stages, compared via F1, $\overline{\log p}_{\text{sent}}$, and JSD. \textsc{Growing} consistently outperforms \textsc{Inward}. 
    For reference,
    the \textsc{Oracle} grammar baseline for $\overline{\log p}_{\text{sent}}=-6.1823$ is provided.
Higher F1 and $\overline{\log p}_{\text{sent}}$, and lower JSD, indicate better performance.
    }
    \label{fig:growingvsinwards}
\end{figure*}

\section{Experimental setup}

\subsection{System}
We use the ``Inside-Outside algorithm for estimating PCFGs from terminal strings" \cite{johnson-2007-doesnt}, which iteratively estimates rule probabilities by optimising likelihood-based objectives from unlabelled strings input. Our setup is based on Mark Johnson's implementation, available on his website\footnote{\url{https://web.science.mq.edu.au/~mjohnson/Software.htm}}. 
This is an example of a grammar reestimation system, which requires as input the symbolic component of a PCFG, i.e. the rules, initialised with a weight and a pseudocount, as well as a list of sentences based on which the probability distributions of the available rules are induced.
We manipulate the pseudocount as discussed in the \textbf{Methodology} section. 
We select to run experiments with 20 iterations because hyperparameter tuning shows plateauing around this point.

\subsection{Data}
\label{sec:data}
We use the child-directed speech sentences and parses from the morphemically tokenised version of CHILDES-TB \cite{pearl-sprouse-2013, marcheva-etal-2025-profiling}, which we restrict to $sentence~length~>1$. The dataset amounts to 126,152 sentences and their corresponding parses. CHILDES-TB\footnote{\url{https://sites.socsci.uci.edu/~lpearl/CoLaLab/CHILDESTreebank/}} is based on the child-directed speech from five corpora, covering 110 children aged from 6 months to 6 years. 

\subsection{\textsc{Oracle} grammar}
We extract a PCFG from the morphemically tokenised CHILDES-TB parses which we use as an \textsc{Oracle} (gold-standard) grammar in evaluation, whilst its symbolic component serves as the initial grammar for the GI system. The \textsc{Oracle} grammar is a PTB-style PCFG approximating adult competence. We do not claim that this PCFG is an absolute representation of adult mental grammar; rather, it provides a linguistically interpretable target against which different staged access hypotheses can be compared.
We restrict production rules by minimum frequency and select the smallest grammar that preserves 100\% parse coverage. At $f_m=7$, the \textsc{Oracle} grammar is comprised of 1,387 production rules and 8,563 lexicalisations over a vocabulary of 6,273 words.

\subsection{Evaluation}
In addition to using the CHILDES-TB parses as a gold standard, based on which we calculate unlabelled F1 score, we also present  distributional metrics which allow for the comparison of the distributions of the learned grammars to the distribution of the \textsc{Oracle} grammar. 

We use an induced grammar to parse 1,000 randomly selected parses from CHILDES-TB. Then we  calculate \textbf{unlabelled F1}, implemented in line with PARSEVAL~\cite{Black1991Coverage}, on parses from the induced grammar and the gold parses from CHILDES-TB.

We evaluate the induced grammars as generative language models using mean length-normalised \textbf{marginal log-likelihood} (see Eq. \ref{eq:loglik}) over 91,901 child speech sentences from the corpora used in CHILDES-TB, restricted to the lexicon of the \textsc{Oracle} grammar.\footnote{A larger child-speech test set could be constructed without this lexical restriction. The age range and number of children match CHILDES-TB.} 
Scores are computed only for utterances licensed by at least one complete parse; larger values are better.
For a PCFG \(G\),
let \(\mathcal{T}(\mathbf{w}^{(i)})\) be the set of all parse trees for an utterance \(\mathbf{w}^{(i)}\):
\begin{equation}
\overline{\log p}_{\text{sent}}= \frac{1}{N} \sum_{i=1}^{N} \frac{\ell_i}{T_i}, \, \, \,  \ell_i = \ln \sum_{t \in \mathcal{T}(\mathbf{w}^{(i)})} P_G(t)
\label{eq:loglik}
\end{equation}


We also compare the rule expansion distributions of \textsc{Oracle} and induced grammars 
for each non-terminal \(A\) using \textbf{Jensen-Shannon divergence (JSD)} \cite{Lin1991}, as formulated in Eq. \ref{eq:JSD}. A result of $JSD=0$ indicates identical distributions.
\begin{equation}
\begin{aligned}
\mathrm{JSD}(p_A, q_A)
&= \tfrac{1}{2}\mathrm{KL}(p_A \| m_A)
 + \tfrac{1}{2}\mathrm{KL}(q_A \| m_A), \\
\text{where}\quad
m_A &= \tfrac{1}{2}(p_A + q_A).
\end{aligned}
\label{eq:JSD}
\end{equation}

\section{Results}
\subsection{Continuity vs. Maturation}

Under the \textsc{Continuity} condition, all grammar rules are available from the outset. In contrast, under \textsc{maturation}, rules are introduced in discrete stages, and the probabilities learned at each stage are carried forward to the next via pseudocount transfer, as defined in Eq.~\ref{eq:existing}. \autoref{tab:f1} provides an overview of the F1 scores achieved by systems approximating the three syntactic development hypotheses we operationalised: \textsc{Continuity}, \textsc{Growing}, \textsc{Inward}.

 \textsc{Continuity} experiments are run for 20 Variational Bayes iterations under a Dirichlet prior with production and lexicalisation pseudocounts $\alpha_p, \alpha_\ell \in \{0.1, 0.3, 0.5\}$. Across these settings, performance is stable: mean F1 (see also \autoref{tab:f1}) is $0.799 \pm 0.001$. 
 We adopt $\alpha_p = \alpha_\ell = 0.1$ as the \textsc{Continuity} baseline for comparison with the \textsc{maturation} conditions, reflecting our choice of minimal constant pseudocounts~(Eq.~\ref{eq:existing}). Under this setting, performance is $F1 = 0.8000$, $\overline{\log p}_{\text{sent}} = -6.0285$, and $JSD = 0.1928$. With appropriate hyperparameter settings, the \textsc{Growing} curriculum outperforms the \textsc{Continuity} condition (see \autoref{tab:f1}), whereas the \textsc{Inward} curriculum consistently underperforms.

\begin{table}[h!]
    \centering
    \footnotesize
    \begin{tabular}{l|l}
        \textbf{Condition} & \textbf{F1} \\
        \hline
        \textsc{Continuity} $\alpha_p, \alpha_\ell \in \{0.1,0.3,0.5\}$ \textit{mean, st.d.} & 0.7994, .0013 \\
        \textsc{Continuity} ($\alpha_p = \alpha_\ell = 0.1$) & 0.8000 \\
        \textsc{Growing} \textit{mean, st.d.} & 0.8059, .0038 \\
        \textsc{Inward} \textit{mean, st.d.} & 0.7953, .0022 \\
        \hline
        \hline
        Neural GI & 0.7856 \\
    \end{tabular}
    \caption{F1 scores comparing \textsc{Continuity} and \textsc{Maturation}. Neural GI baseline from Table~2 in Marcheva et al (2025) is provided for comparison.}
    \label{tab:f1}
\end{table}

\subsection{\textsc{Growing} vs. \textsc{Inward} }
\subsubsection{Final grammar comparison}
We compare final grammars, after stage 5, induced under the \textsc{Growing} and \textsc{\textsc{Inward}} curricula. We report the top 10 hyperparameter settings (ranked by \textsc{\textsc{Growing}} F1) in \autoref{tab:inwards_vs_growing_selected}; in all cases, \textsc{\textsc{Growing}} also exceeds the \textsc{Continuity} baseline ($\alpha_p=\alpha_\ell=0.1$, $F1=0.8000$).
Paired, one-sided Wilcoxon signed-rank tests over 72 matched hyperparameter configurations spanning $s_\ell \in \{0.001,0.01,0.1\}$, $s_p \in \{0.001,0.005,0.01,0.05,0.1,0.2\}$, and $\eta \in \{0.001,0.005,0.01,0.05\}$ show that \textsc{\textsc{Growing}} significantly outperforms \textsc{\textsc{Inward}} on all evaluation metrics (see Table~\ref{tab:growing_vs_inwards_stats}). 
The underperformance of \textsc{Inward} may be explained either by the early introduction of INTJ, or by the very late introduction of VP. As observed in  \autoref{fig:growingvsinwards}, the final stage for \textsc{Inward}, corresponding to the VP stage in \autoref{tab:syntactic-stages}, leads to the largest improvement in performance across all three metrics. This makes it more likely that the underperformance of \textsc{Inward} is due to the late introduction of VP. 
This implies that \textsc{Growing} better captures utterances requiring predicate, argument, and modifier structure, whereas any \textsc{Inward} advantage is restricted to interactional or clause-level material. 

\begin{table}[h!]
\centering
\footnotesize
\begin{tabular}{lcccc}
Metric & Median $\Delta$   & $p$-value & $r_{\text{rank-biserial}}$ & Direction \\
\hline
F1 & $+0.0056$  & $1.5\times10^{-12}$ & $0.95$ & \textsc{Growing} \\
 $\overline{\log p}_{\text{sent}}$  & $+0.0104$  & $7.4\times10^{-13}$ & $0.96$ & \textsc{Growing} \\
JSD  & $-0.0198$  & $8.3\times10^{-14}$ & $-1.00$ & \textsc{Growing}\\
\end{tabular}
\caption{Results from a paired one-sided Wilcoxon signed-rank test comparing 72 pairs of final grammars induced under \textsc{\textsc{Growing}} and \textsc{\textsc{Inward}}. }
\label{tab:growing_vs_inwards_stats}
\end{table}


\begin{table}[t]
\small
\centering
\setlength{\tabcolsep}{2pt}

\begin{tabular}{cc|ccc|ccc}
\hline
& &
\multicolumn{3}{c|}{\textbf{\textsc{Growing}}} &
\multicolumn{3}{c}{\textbf{\textsc{Inward}}} \\
$s_p$ & $\eta$ &
F1 & $\overline{\log p}_{\text{sent}}$ & JSD &
F1 & $\overline{\log p}_{\text{sent}}$ & JSD \\
\hline
0.01  & 0.001 & 0.8121 & -6.0545 & 0.2558 & 0.7935 & -6.0941 & 0.2739 \\
0.01  & 0.05  & 0.8106 & -6.0542 & 0.2548 & 0.7943 & -6.0941 & 0.2727 \\
0.01  & 0.01  & 0.8103 & -6.0543 & 0.2555 & 0.7943 & -6.0941 & 0.2736 \\
0.01  & 0.005 & 0.8102 & -6.0544 & 0.2557 & 0.7936 & -6.0941 & 0.2738 \\
0.005 & 0.005 & 0.8080 & -6.0547 & 0.2572 & 0.7927 & -6.0929 & 0.2730 \\
0.005 & 0.01  & 0.8078 & -6.0546 & 0.2572 & 0.7929 & -6.0929 & 0.2730 \\
0.001 & 0.001 & 0.8077 & -6.0543 & 0.2584 & 0.7976 & -6.0908 & 0.2726 \\
0.001 & 0.01  & 0.8070 & -6.0543 & 0.2583 & 0.7978 & -6.0908 & 0.2726 \\
0.005 & 0.001 & 0.8070 & -6.0547 & 0.2573 & 0.7929 & -6.0929 & 0.2730 \\
0.001 & 0.005 & 0.8068 & -6.0543 & 0.2584 & 0.7974 & -6.0908 & 0.2726 \\
\hline
\end{tabular}
\caption{Top-performing hyperparameter setups for \textbf{\textsc{Growing}} and \textbf{\textsc{Inward}} curricula.
$s_\ell=0.1$ for all of these. Higher F1 and $\overline{\log p}_{\text{sent}}$, and lower JSD, indicate better performance.}
\label{tab:inwards_vs_growing_selected}
\end{table}

\subsubsection{Learning progression}
The learning progression through the stages of the two maturational curricula is illustrated in \autoref{fig:growingvsinwards}. The F1 under \textsc{Growing} is consistently higher throughout the stages. Furthermore, the learning progression of the \textsc{Growing} curriculum for all metrics indicates monotonic improvement, in contrast with \textsc{Inward} where some stages decrease performance (e.g. stage 1-2 for $\overline{\log p}_{\text{sent}}$ or stage 2-3 for mean JSD). For  \textsc{Inward}'s F1 score, the last stage is primarily responsible for the final F1 achieved. This corresponds to the VP stage, which introduces argument movement. The mean JSD reflects a similar pattern to F1.
The plot of the mean normalised sentence log-likelihood, $\overline{\log p}_{\text{sent}}$, shows that both maturational conditions lead to improvement over the oracle (initial grammar). This result affirms that the maturational hypotheses produce grammars more favourable to explaining child productions.

\subsubsection{JSD per phrase NT}
\begin{figure*}[t!]
    \centering
    \includegraphics[width=\linewidth]{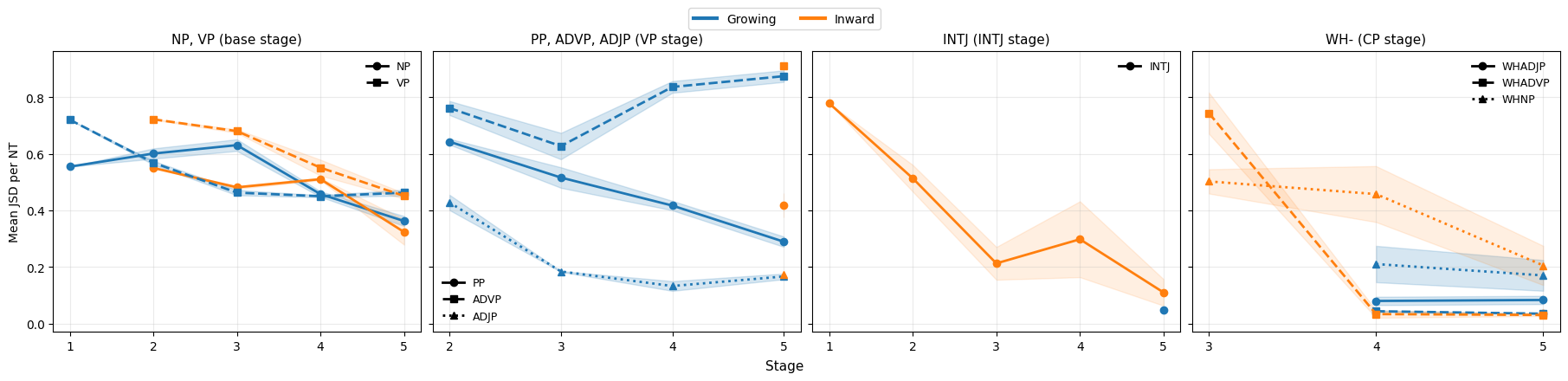}
    \caption{Mean JSD for instrumental phrase-level NTs. \textsc{Growing} is in blue and \textsc{Inward} is in orange, but the different NTs are illustrated with different figure markers and line styles. Low scores of JSD indicate similarity with the \textsc{Oracle} distribution. }
    \label{fig:individual_jsd}
\end{figure*}

We examine the stage-wise progression of JSD for selected phrase-level NTs introduced at different points under the \textsc{Growing} and \textsc{Inward} curricula. Tracking individual categories allows us to compare how alternative maturational orderings affect the stabilisation of specific components of the grammar. 

Under \textsc{Growing}, core phrasal categories such as NP and VP show gradual, largely monotonic reductions in JSD, with a similar pattern for VP-stage modifier phrases (PP, ADVP, ADJP). Under \textsc{Inward}, these VP-related categories, which are introduced only at the final stage, exhibit higher JSD relative to \textsc{Growing}, suggesting as expected, that the more stages where the category is active would lead to a more refined result. For clause-level categories, including INTJ and WH-related phrases, the two curricula show more mixed trajectories, with no uniform advantage for either ordering. 
By making the emergence of NTs explicit, we provide a quantitative tool for exploring how different developmental orderings shape learning dynamics.

\section{Discussion}
While our experiments showcase a flexible computational framework for comparing syntactic development theories under controlled conditions, several limitations of the present study should be acknowledged. First, our evaluation relies on CHILDES-TB, which provides manually annotated parses for English CDS. Although the GI framework itself is not language-specific, extending this approach to other languages requires the appropriately annotated resources. Second, the curricula may differ in the number of rules available at each developmental stage. Future work could further control for this factor by testing more fine-grained curricula. Third, the type of GI our framework relies on is grammar reestimation. The symbolic part of the grammar needs to be provided, and learning consists of estimating the probabilistic component of the grammar from the input, which consists of unlabelled sentences. This design choice ensures a controlled comparison between maturational hypotheses but limits the model's ability to discover novel structures.


Although we operationalise two maturational hypotheses, \textsc{Growing} and \textsc{Inward}, our pipeline for staged syntactic development, illustrated in \autoref{fig:pipeline} is not tied to maturation as the source for ordering in syntactic development. 
Usage-based and constructivist theories also reject continuity, i.e. that children have adult grammatical knowledge from the outset \cite{Tomasello2000, TOMASELLO2005, BEHRENS2021}. 
The main difference between maturational and usage-based accounts is the source of the order of syntactic development: under maturation, the order of categories is innate, whereas under usage-based accounts it emerges constrained by the input and domain-general cognitive capacities of the child.
Our pipeline requires that the stages of syntactic development are pre-specified before training. Thus, it can be used for non-generativist accounts, if a curriculum is specified based on item-specific frames or constructions. In that way, our staged 
grammar-induction paradigm provides a general method for comparing 
theories that reject the \textsc{Continuity} hypothesis, even if they disagree about why the learner's hypothesis space changes over time.

\section{Conclusion}
We present a grammar induction framework for staged syntactic development hypotheses, which makes the emergence of syntactic categories explicit and enables a controlled comparison of alternative developmental orderings. 
We operationalise two competing maturational hypotheses of syntactic development, \textsc{Growing} and \textsc{Inward}, and find that 
under our framework \textsc{Growing} significantly outperforms \textsc{Inward}.
This pilot comparison
illustrates how our framework can serve as a principled tool for investigating alternative staged hypotheses of syntactic development.


\bibliographystyle{apacite}

\setlength{\bibleftmargin}{.125in}
\setlength{\bibindent}{-\bibleftmargin}

\bibliography{references}

\end{document}